\documentclass[10pt, a4paper]{article}

\usepackage{lrec2026} 
\usepackage{tikz}
\usetikzlibrary{positioning,arrows.meta,calc}
\usepackage{graphicx} 
\usepackage{linguex}
\usepackage[table]{xcolor}
\usepackage{booktabs}
\usepackage{enumitem}
\usepackage[most]{tcolorbox}
\usepackage{placeins}
\usepackage{pgfplots}
\usepackage{hyperref}
\usepackage{booktabs}
\usepackage{array}
\usepackage{dblfloatfix}
\usepackage{float}

\newcommand{\DI}{\textit{-DI}}
\newcommand{\MIS}{\textit{-mI\c{s}}}
\newcommand{\model}[1]{\texttt{\detokenize{#1}}}

\pgfplotsset{compat=1.18} 
\newtcblisting{PromptBox}[1]{%
  breakable,
  colback=gray!5,
  colframe=black!40,
  title={#1},
  listing only,
  listing options={
    basicstyle=\ttfamily\footnotesize,
    breaklines=true,
    breakatwhitespace=true
  }
}

\definecolor{EviRoot}{RGB}{232,240,254}   
\definecolor{EviDirect}{RGB}{232,245,233} 
\definecolor{EviIndirect}{RGB}{255,243,224} 
\definecolor{EviSubA}{RGB}{243,248,255}   
\definecolor{EviSubB}{RGB}{255,249,240}   
\definecolor{EviNote}{RGB}{245,245,245}   

\title{Benchmarking Source-Sensitive Reasoning in Turkish:\\
Humans and LLMs under Evidential Trust Manipulation}

\name{Sercan Karaka\c{s}$^1$, Yusuf \c{S}im\c{s}ek$^2$}

\address{$^1$University of Chicago, skarakas@uchicago.edu \\
         $^2$F{\i}rat University, ysimsek@firat.edu.tr\\}

\abstract{
This paper investigates whether \emph{source trustworthiness} shapes Turkish evidential morphology and whether large language models (LLMs) track this sensitivity. We study the past-domain contrast between \textit{-DI} and \textit{-mI\c{s}} in controlled cloze contexts where the information source is overtly external, while only its perceived reliability is manipulated (High-Trust vs.\ Low-Trust). In a human production experiment, native speakers of Turkish show a robust trust effect: High-Trust contexts yield relatively more \textit{-DI}, whereas Low-Trust contexts yield relatively more \textit{-mI\c{s}}, with the pattern remaining stable across sensitivity analyses. We then evaluate 10 LLMs in three prompting paradigms (open gap-fill, explicit past-tense gap-fill, and forced-choice A/B selection). LLM behavior is highly model- and prompt-dependent: some models show weak or local trust-consistent shifts, but effects are generally unstable, often reversed, and frequently overshadowed by output-compliance problems and strong base-rate suffix preferences. The results provide new evidence for a trust-/commitment-based account of Turkish evidentiality and reveal a clear human--LLM gap in source-sensitive evidential reasoning.
\\ \newline 
}

\begin{document}
\pagestyle{plain}
\pagenumbering{arabic}
\setcounter{page}{1}

\maketitleabstract

\section{Introduction}

Evidentiality refers to the linguistic encoding of information source, that is, how a speaker indicates the basis on which a proposition is presented (e.g., direct perception, inference, or report) \cite{willett1988evidentiality,dendaletasmowski2001,dehaan2001inference,plungian2001evidentiality,aikhenvald2004evidentiality,boye2012epistemic,unalpapafragou2020}. In many languages, evidential markers systematically signal whether the speaker witnessed an event, inferred it from available evidence, or learned it from someone else. Because evidentiality lies at the interface of semantics, pragmatics, and discourse, it has been central to research on speaker commitment, reliability, and perspective in natural language.

A useful perspective on evidentiality comes from language acquisition and “thinking for speaking”: as children learn a language, they learn to attend to distinctions that the language regularly encodes \cite{brownlenneberg1954,slobin1996}. In evidential languages, this can make knowledge source (e.g., direct experience, inference, report) especially salient early on, since speakers must repeatedly track and express it in discourse. For Turkish, this suggests that evidential morphology is not just grammatical, but part of a broader cognitive routine linking source monitoring, memory, and speaker stance \cite{aksukocslobin1986}. Turkish offers a particularly important case for evidentiality research because past-domain morphology is closely tied to distinctions in information source and epistemic stance \cite{aksukocslobin1986,underhill1976,johanson2003,gokselkerslake2005,kornfilt1997}. A well-known contrast is between the suffix \textit{-DI}, traditionally associated with direct evidence or stronger speaker commitment, and the suffix \textit{-mI\c{s}}, which is commonly associated with indirect evidence, such as inference or report \cite{aksukocslobin1986,underhill1976,johanson2003,gul2008,sener2011,gokselkerslake2005}.

\begin{figure}[t]
\centering
\resizebox{\columnwidth}{!}{%
\begin{tikzpicture}[
  node distance=5mm and 7mm,
  rootbox/.style={
    draw=blue!50!black,
    fill=EviRoot,
    rounded corners,
    align=center,
    inner sep=2.5pt,
    font=\footnotesize
  },
  dbox/.style={
    draw=green!45!black,
    fill=EviDirect,
    rounded corners,
    align=center,
    inner sep=2.5pt,
    font=\footnotesize
  },
  ibox/.style={
    draw=orange!60!black,
    fill=EviIndirect,
    rounded corners,
    align=center,
    inner sep=2.5pt,
    font=\footnotesize
  },
  sboxa/.style={
    draw=blue!40!black,
    fill=EviSubA,
    rounded corners,
    align=center,
    inner sep=2pt,
    font=\scriptsize
  },
  sboxb/.style={
    draw=orange!50!black,
    fill=EviSubB,
    rounded corners,
    align=center,
    inner sep=2pt,
    font=\scriptsize
  },
  notebox/.style={
    draw=black!50,
    fill=EviNote,
    rounded corners,
    align=center,
    inner sep=2pt,
    font=\scriptsize
  },
  line/.style={
    -Latex,
    semithick,
    draw=black!75
  }
]

\node[rootbox] (ev) {\textbf{Evidentiality}\\[-0.5mm]\scriptsize marking information source};

\node[dbox, below left=7mm and 10mm of ev] (direct)
  {\textbf{Direct}\\\scriptsize firsthand access};
\node[ibox, below right=7mm and 10mm of ev] (indirect)
  {\textbf{Indirect}\\\scriptsize non-firsthand access};

\draw[line] (ev) -- (direct);
\draw[line] (ev) -- (indirect);

\node[sboxa, below left=6mm and 4mm of direct] (visual) {Visual};
\node[sboxa, below right=6mm and 4mm of direct] (nonvisual) {Non-visual sensory};

\draw[line] (direct) -- (visual);
\draw[line] (direct) -- (nonvisual);

\node[sboxb, below left=6mm and 4mm of indirect] (infer) {Inferential};
\node[sboxb, below right=6mm and 4mm of indirect] (report) {Reported / hearsay};

\draw[line] (indirect) -- (infer);
\draw[line] (indirect) -- (report);

\node[notebox, below=8mm of $(visual)!0.5!(report)$, text width=6.8cm] (note) { languages vary in how they grammaticalize these distinctions,
  and many systems do not map neatly onto a single taxonomy.
};

\draw[dashed, draw=black!60] (ev.south) -- (note.north);

\end{tikzpicture}%
}
\caption{Information-source marking, adapted from typological overviews of evidential systems \cite{willett1988evidentiality,aikhenvald2004evidentiality}.}
\label{fig:evidentiality-general}
\end{figure}
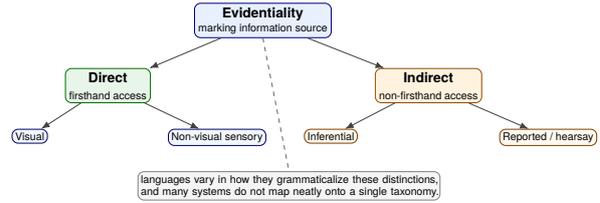

A minimal illustration is given in \ref{ex:di}--\ref{ex:mis}: in \ref{ex:di}, the use of \textit{-DI} is classically associated with a context in which the speaker directly witnessed the event, whereas in \ref{ex:mis}, the use of \textit{-mI\c{s}} suggests that the speaker did not directly witness the event but instead inferred it or learned it from another source. At the same time, a substantial body of work notes that the actual distribution and interpretation of these forms are sensitive to discourse context, information structure, and speaker stance, so the traditional labels are best understood as useful descriptive generalizations rather than rigid one-to-one mappings between form and meaning \cite{gul2008,sener2011}.

{\footnotesize
\setlength{\Exlabelsep}{0.5em}

\ex.\label{ex:di}
\a. \gll Ahmet Ali'yi gör-dü.\\
        Ahmet Ali-\textsc{acc} see-\textsc{pst.dir}\\
    \glt `Ahmet saw Ali (witnessed/direct evidence).'
\z.

\ex.\label{ex:mis}
\a. \gll Ahmet Ali'yi gör-mü\c{s}.\\
        Ahmet Ali-\textsc{acc} see-\textsc{pst.indir}\\
     \glt `Ahmet apparently/reportedly saw Ali (hearsay/inference).'
\z.

}

The \textit{-DI}/\textit{-mI\c{s}} contrast offers a well-studied case for investigating how languages encode information source, speaker stance, and pragmatic inference \cite{aksukocslobin1986,johanson2003,gokselkerslake2005,gul2008,sener2011,kandemirci2023,atamandevrim2025}, and it is also central to developmental work on how children learn to track and express knowledge source in discourse \cite{aksukocslobin1986,slobin1996}. In addition, evidential distinctions are highly relevant for NLP tasks such as translation, summarization, and source-sensitive interpretation.

\subsection{Why evidentiality matters for LLM research?}

Evidentiality is also increasingly relevant to LLM research, because many of the core reliability problems in modern text generation are, in effect, problems about \emph{source status}: whether a claim is grounded in direct input evidence, retrieved documents, parametric memory, inference, or unsupported generation. Recent work on factuality and hallucination has shown that fluent model outputs may be unfaithful to available evidence or false in world knowledge terms, motivating finer-grained analyses of how models represent and communicate the basis of their claims \citep{maynez-etal-2020-faithfulness,lin-etal-2022-truthfulqa,manakul-etal-2023-selfcheckgpt,min-etal-2023-factscore,bang-etal-2025-hallulens,liu-etal-2026-agenthallu}. In parallel, work on calibration and self-evaluation suggests that models can sometimes track uncertainty or answerability, but this ability is uneven and sensitive to task format and available context \citep{kadavath2022lmmostly}. At the same time, frontier and open models are improving rapidly in reasoning, long-context processing, multilinguality, and agentic performance, as reflected in recent reports  \citep{singh2025openaigpt5card,yang2025qwen3technicalreport}. These gains make source-sensitive evaluation more, not less, important: as models become stronger and more fluent, the central question is increasingly not only whether they can produce plausible language, but whether they can reliably represent the evidential basis of what they say.

From this perspective, evidentiality offers a linguistically grounded framework for studying how systems encode (or fail to encode) distinctions among \emph{witnessed}, \emph{inferred}, and \emph{reported} information. This is especially important in settings that explicitly require provenance, such as retrieval-augmented generation and citation-based QA, where models must connect claims to supporting evidence and where attribution quality itself has become a major evaluation target \citep{lewis2020rag,nakano2022webgpt,gao-etal-2023-enabling,li-etal-2024-attributionbench}. More broadly, evidential categories provide a principled way to ask whether LLMs merely optimize surface plausibility or whether they track source-sensitive epistemic distinctions that are central to human communication. Besides, this line of work is also important because direct \emph{human--LLM} comparisons for Turkish remain relatively scarce, especially for source-sensitive semantic/pragmatic phenomena such as evidentiality. Much of the recent Turkish LLM literature has focused on model development, benchmarking, and engineering issues \citep{bayram2025tokenizatio,bayram2025tokenization,er2025cetvel,isbarov-etal-2025-tumlu,umutlu-etal-2025-evaluating,wicaksono2025emotiontr,bayram2026tokensmeaninghybridtokenization,karakas2026lemmas,toraman2026turkbench,ugur2026mecellem}, rather than tightly controlled comparisons between human judgments and model preferences on linguistically targeted contrasts. Emerging Turkish-focused human--LLM studies suggest that such comparisons can reveal substantial mismatches between surface-form preferences and human-like contextual reasoning in several linguistic domains \citep{keles-deniz-2024-superficial,keles-deniz-2025-men,karakas2026turkishreflexivebinding,karakas2026plausibility}.

\section{Related Work}
\label{sec:related-work}

\paragraph{Turkish evidentiality: traditional descriptive and grammatical accounts.}
Turkish has long occupied a central place in the evidentiality literature because its past-domain morphology is closely tied to distinctions in information source and speaker stance \citep{underhill1976,aksukocslobin1986,johanson2003,kornfilt1997,gokselkerslake2005}. A traditional descriptive contrast distinguishes \textit{-DI} and \textit{-mI\c{s}}: \textit{-DI} is often introduced as the form associated with direct evidence (or stronger speaker commitment), while \textit{-mI\c{s}} is associated with indirect evidence, especially inferential and reportative uses \citep{underhill1976,johanson2003,gokselkerslake2005,aksukocslobin1986}. This contrast has also been influential in acquisition and discourse-based work, where Turkish evidential morphology is treated as a system that helps speakers track and communicate how information was obtained \citep{aksukocslobin1986}.

At the same time, the Turkish literature shows that the traditional mapping is not a simple one-to-one encoding. A central debate is whether \textit{-DI} is a true direct evidential, or primarily a past tense form whose ``directness'' and stronger commitment effects arise through discourse-pragmatic inference in canonical contexts \citep{gul2008,sener2011}. Likewise, analyses of \textit{-mI\c{s}} emphasize sensitivity to evidential source, discourse configuration, and speaker perspective, rather than a single fixed meaning \citep{gul2008,johanson2003}. For this reason, recent work often treats \textit{direct}/\textit{indirect} as useful descriptive labels while allowing context-sensitive meaning and pragmatic strengthening.

For the present study, we adopt Giannakidou and Mari’s framework on veridicality, nonveridicality, commitment, and evidential bias \citep{giannakidou1998,giannakidou2006,giannakidoumari2016,giannakidoumari2018,giannakidoumari2021}. This framework is useful because it separates notions often conflated in evidentiality research: (i) information source, (ii) speaker commitment, and (iii) the structure of the speaker’s information state. It also motivates treating source trustworthiness as a factor shaping evidential interpretation and commitment \citep{boscaro-giannakidou-mari-2025}.

Let $M=M_i(w,t)$ be speaker $i$'s information state at world $w$ and time $t$. A compact way to encode the relevant distinction is to contrast veridical and nonveridical states:

\ex.\label{ex:veridical-state}
$\textsc{Veridical}_M(p)$ iff all worlds in $M$ are $p$-worlds:
$\forall w \in M\,[p(w)]$.

\ex.\label{ex:nonveridical-state}
$\textsc{Nonveridical}_M(p)$ iff $M$ contains both a $p$-world and a $\neg p$-world:
$\exists w \in M\,[p(w)]$ and $\exists v \in M\,[\neg p(v)]$.

The formula in~\ref{ex:veridical-state} states that the speaker's information state supports only $p$-worlds (full support for $p$), while the formula in~\ref{ex:nonveridical-state} states that the information state remains compatible with both $p$ and $\neg p$ possibilities. In this perspective, evidential and modal expressions can encode an evidence-based bias toward $p$ without requiring full veridical commitment.

This distinction is useful for Turkish evidentiality because it allows us to ask not only which source type is invoked, but also whether that source is treated as strong enough to support a more veridical commitment profile. In our trust-based design, source trustworthiness is therefore modeled as a factor that can shift speakers between stronger commitment and weaker, evidence-biased interpretations, rather than as a purely descriptive source label.

\section{Methods}
\label{sec:methods}

To examine how Turkish evidential morphology responds to contextual
\emph{trustworthiness}, we used two complementary experimental paradigms with
two corresponding datasets. First, we designed a controlled cloze-style task
that can be administered to both human participants and LLMs. Second, we
constructed a larger dataset for LLM-only evaluation in order to test
generalization over a broader set of contexts. In all tasks, the target is a
\emph{cloze} (fill-in-the-blank) completion: participants or models infer and
produce the missing final word of a sentence from the preceding context. This
shared format provides a direct measure of context-sensitive morphological
preferences in production and enables human--model comparison within a single
task setup.

\subsection{Dataset 1 for Human and LLM Comparison}
Dataset 1 consists of 60 original items, partitioned into three categories: high-trust, low-trust, and filler sentences. All items were manually authored by the researchers. High-trust sources correspond to announcement channels that are generally considered reliable in everyday life, whereas low-trust sources typically consist of channels that are difficult to trust. These sources are provided in Appendix~\ref{app:source_categories}. During construction, we aimed to make the target completion
as uniquely and unambiguously recoverable as possible, thereby minimizing cases
with multiple equally plausible completions, while also balancing potential
confounds such as lexical frequency and sentence length across conditions. In
the cloze task, participants were asked to complete the final blank with the
most natural and contextually appropriate continuation.

{\footnotesize
\ex.\label{ex:hightrust}
\a. Belediyenin SMS uyarısına göre sular .......\\
    `According to the municipality's SMS alert, the water ........'
\z.

\ex.\label{ex:lowtrust}
\a. Yan binadaki teyzenin dediğine göre sular .......\\
    `According to the lady next door, the water .......'
\z.

}

Across conditions, the source of information is kept \emph{overtly external}
through explicit attribution frames (e.g., \emph{X'e g\"ore} `according to X';
cf.\ \ref{ex:hightrust}--\ref{ex:lowtrust}). The manipulation targets the
\emph{perceived reliability} of that external source: High-Trust items present
the proposition as coming from institutionally authoritative channels (e.g., an official municipal SMS notification), whereas Low-Trust items present it as coming from
informal or weakly accountable channels (e.g., a statement by a neighbor). To verify that this manipulation isolates trust rather than
idiosyncratic item properties, we ran a separate norming study with a broad
participant pool. Participants consistently rated High-Trust sources as more
credible than Low-Trust sources, and items that failed to show the intended
separation were revised or removed.

\subsection{Dataset 2 for LLM-Only Evaluation}
Dataset~2 is an expanded item set containing all experimental items from Dataset~1 together with additional items, and was designed exclusively for large language model evaluation. It preserves the same structural properties and manipulation logic as Dataset~1, while increasing coverage and statistical stability by substantially expanding the number of scenarios and lexical realizations. The expanded set
contains 200 items in total. Dataset~2 was \emph{not} administered to human
participants; instead, it was designed to test model behavior on a broader item
base without increasing participant burden, and to assess whether patterns
observed in the human-scale dataset replicate and generalize when the trust
manipulation is instantiated across a wider range of contexts.

\section{Experiments}
\label{sec:experiments}

\subsection{Human Experiment}
\label{sec:human}

The human experiment received prior approval from the University of Chicago Institutional Review Board (IRB26-0198). All participants provided informed consent before participation. We recruited 75 unique participants who self-identified as native speakers of Turkish, were at least 18 years old, and were residing in Turkey at the time of participation. The experiment was implemented and administered online using the \textsc{PCIbex}/\textsc{PennController} platform \citep{zehrschwarz2018penncontroller}. The main study produced 4{,}500 trials in total: 1{,}500 High-Trust trials, 1{,}500 Low-Trust trials, and 1{,}500 filler trials. For each trial, we recorded the participant's completion response. Our primary analyses target the critical High-Trust vs.\ Low-Trust contrast, so we analyze 3{,}000 trials after excluding fillers. All responses were analyzed in de-identified form. This balanced design allows a controlled test of how the trust manipulation affects production preferences.

\subsection{LLM Experiments}
\label{sec:llm}
In this study, LLMs were tested using two different datasets and ten models \citep{gemmateam2025gemma3report, gemma2tr, novus7b, orbita, yang2025qwen3technicalreport, trendyol7b, turkcell7b, bezir2024wiroaiturkishllm, gptoss20b, gpt2tr10m}. and three prompts.The models evaluated in this study consist of Turkish-focused and multilingual large language models. gpt2-turkish-10M, WiroAI-turkish-LLM-8B, Turkcell-LLM-7B, Trendyol-LLM-7B, Gemma-2-9B-IT-TR, and Novus-7B-TR are models that have been either directly trained on Turkish or fine-tuned with Turkish data. In contrast, Gemma-3 and Qwen3-32B are trained on large-scale multilingual data and are capable of processing Turkish through their multilingual capacities. Some models, on the other hand, are built on a multilingual foundation and later fine-tuned with Turkish instruction data. This diversity allows us to compare the effects of Turkish-focused and multilingual training approaches on sensitivity to evidentiality morphology. During the study, the models were asked to fill in the blank format that was created \ref{ex:hightrust}--\ref{ex:lowtrust}. Three prompts were used to perform this task; these prompts are given in Appendix \ref{app:prompts}. While the models were expected to generate the word in the space indicated by the space in the first two prompts, in the third prompt two different words were given as options, one of which is the word conjugated with "-DI" \ref{ex:hightrust_choose}, and the other is the word conjugated with "-mIş" \ref{ex:lowtrust_choose}. In the third prompt, the models were expected to choose the most appropriate word between these two options according to the flow of the sentence. While generating the model outputs, the temperature was set to 0.1 and the top-p to 1; an additional max token setting of 8 was added to the third prompt.

{\footnotesize
\ex.\phantomsection\label{ex:hightrust_choose}
\a. \gll Getir-il-di \\
        bring-\textsc{pass}-\textsc{pst.dir} \\
    \glt `was brought (-DI form)'
\z.

\ex.\phantomsection\label{ex:lowtrust_choose}
\a. \gll Getir-il-mi\c{s} \\
        bring-\textsc{pass}-\textsc{pst.indir} \\
    \glt `apparently/ reportedly was brought (-mIş form)'
\z.

}

\section{Human experiment results}
\label{sec:human-results}

\subsection{Production: trust robustly shifts \textit{-DI} vs.\ \textit{-mI\c{s}}}
\label{sec:human-prod}

Table~\ref{tab:strict-counts} shows strict-coding counts for critical trials
(High+Low; fillers excluded). Descriptively, High-trust contexts produce more
\textsc{DI} completions, whereas Low-trust contexts produce more
\textsc{mI\c{s}} completions. The \textsc{Other} category is similar across
conditions (High: 28.4\%; Low: 29.5\%), suggesting that the trust manipulation
primarily redistributes responses \emph{within} the evidential morphology space
rather than increasing off-target productions.

\definecolor{tblheader}{RGB}{235,239,245}
\definecolor{hightrustrow}{RGB}{232,244,255}
\definecolor{lowtrustrow}{RGB}{255,241,232}
\definecolor{dihi}{RGB}{214,236,255}     
\definecolor{mislo}{RGB}{255,228,210}    

\begin{table}[!htbp]
\centering
\small
\resizebox{0.9\columnwidth}{!}{%
\setlength{\tabcolsep}{4pt}
\renewcommand{\arraystretch}{1.05}
\begin{tabular}{lccc}
\toprule
\rowcolor{tblheader}
\textbf{Condition} & \textbf{\textsc{DI}} & \textbf{\textsc{mI\c{s}}} & \textbf{Other} \\
\midrule
\rowcolor{hightrustrow}
High trust & \cellcolor{dihi}\textbf{727} (48.5\%) & 347 (23.1\%) & 426 (28.4\%) \\
\rowcolor{lowtrustrow}
Low trust  & 548 (36.5\%) & \cellcolor{mislo}\textbf{509} (33.9\%) & 443 (29.5\%) \\
\bottomrule
\end{tabular}
}
\caption{Strict coding counts by condition (critical trials only; $n=1500$ per condition). Percentages are within condition.}
\label{tab:strict-counts}
\end{table}

\begin{figure}[!htbp]
\centering
\resizebox{0.80\columnwidth}{!}{%
\begin{minipage}{0.48\linewidth}
\centering
\begin{tikzpicture}
\begin{axis}[
  ybar,
  bar width=10pt,
  ymin=0, ymax=1,
  ylabel={Proportion},
  symbolic x coords={DI,mI\c{s}},
  xtick=data,
  ymajorgrids=true,
  width=\linewidth,
  height=3.6cm,
  title={High trust}
]
\addplot coordinates {(DI,0.677) (mI\c{s},0.323)};
\end{axis}
\end{tikzpicture}
\end{minipage}\hfill
\begin{minipage}{0.48\linewidth}
\centering
\begin{tikzpicture}
\begin{axis}[
  ybar,
  bar width=10pt,
  ymin=0, ymax=1,
  ylabel={Proportion},
  symbolic x coords={DI,mI\c{s}},
  xtick=data,
  ymajorgrids=true,
  width=\linewidth,
  height=3.6cm,
  title={Low trust}
]
\addplot coordinates {(DI,0.518) (mI\c{s},0.482)};
\end{axis}
\end{tikzpicture}
\end{minipage}%
}

\caption{Human production (strict coding): within-condition proportions of \textit{-DI} vs.\ \textit{-mI\c{s}} among DI/mI\c{s} responses only.}
\label{fig:di-mis-by-trust}
\end{figure}
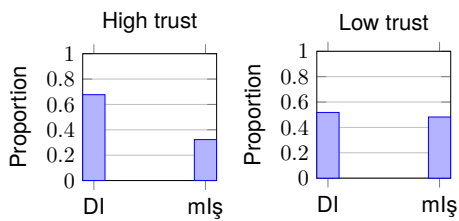

Figure~\ref{fig:di-mis-by-trust} plots the \emph{within-condition} proportions
of \textsc{DI} vs.\ \textsc{mI\c{s}} after excluding \textsc{Other}. In the
High-trust condition, $\textsc{DI}=727/(727+347)=0.677$ and
$\textsc{mI\c{s}}=0.323$. In the Low-trust condition,
$\textsc{DI}=548/(548+509)=0.518$ and $\textsc{mI\c{s}}=0.482$.

Two robustness checks yield the same qualitative pattern. (i) Under lenient
last-token coding (allowing two-word VP completions), the association remains
strong ($\chi^2(1)=52.685$, $p=3.92\times 10^{-13}$; OR $=1.859$, 95\% CI
$[1.571,\,2.199]$). (ii) A content-controlled stratified analysis with matched
content keys gives a pooled OR $=2.029$, 95\% CI $[1.654,\,2.490]$ (CMH $\chi^2(1)=46.789$, $p=7.91\times 10^{-12}$), with no strong evidence of heterogeneity across strata ($p=0.137$).

\begin{table*}[t]
\centering
\scriptsize
\setlength{\tabcolsep}{3.5pt}
\renewcommand{\arraystretch}{1.05}
\resizebox{\textwidth}{!}{%
\begin{tabular}{lrrrrrrrrr}
\toprule
\textbf{Model} &
\textbf{Usable} &
\textbf{H-DI} &
\textbf{H-MI\c{S}} &
\textbf{L-DI} &
\textbf{L-MI\c{S}} &
\textbf{DI\% (H)} &
\textbf{DI\% (L)} &
\textbf{$\Delta$DI (pp)} &
\textbf{$p$} \\
\midrule
Gemma-3-27B-IT        & 64.5 & 195 & 3  & 168 & 21 & 98.5 & 88.9 & +9.6  & $7.44\times10^{-5}$ \\
Gemma-2-9B-IT-TR      & 47.2 & 158 & 2  & 113 & 10 & 98.8 & 91.9 & +6.9  & $0.006$ \\
Orbita-v0.1           & 40.3 & 83  & 40 & 97  & 22 & 67.5 & 81.5 & -14.0 & $0.018$ \\
Qwen3-32B             & 24.7 & 54  & 3  & 81  & 10 & 94.7 & 89.0 & +5.7  & $0.371$ \\
Novus-7B-TR           & 21.3 & 79  & 1  & 46  & 2  & 98.8 & 95.8 & +2.9  & $0.556$ \\
WiroAI-TR-8B          & 16.8 & 54  & 5  & 40  & 2  & 91.5 & 95.2 & -3.7  & $0.696$ \\
GPT-OSS-20B-Astro     & 11.3 & 29  & 11 & 26  & 2  & 72.5 & 92.9 & -20.4 & $0.058$ \\
GPT2-TR-10M           & 2.8  & 10  & 0  & 7   & 0  & 100.0 & 100.0 & +0.0  & -- \\
Turkcell-LLM-7B       & 1.2  & 5   & 0  & 2   & 0  & 100.0 & 100.0 & +0.0  & -- \\
Trendyol-LLM-7B       & 0.0  & 0   & 0  & 0   & 0  & --    & --    & --    & -- \\
\bottomrule
\end{tabular}%
}
\caption{Experiment~I (prompting-based gap-fill) LLM results. \textbf{Usable} = percentage of all generations classifiable as \textsc{DI} or \textsc{mI\c{s}} (i.e., DI/MI\c{S}\%). DI\% is computed as $\mathrm{DI}/(\mathrm{DI}+\mathrm{MIS})$ within each trust condition. $\Delta$DI is High minus Low (percentage points). $p$ values are Fisher exact tests on \{High, Low\}$\times$\{\textsc{DI}, \textsc{mI\c{s}}\}.}
\label{tab:llm-exp1-main}
\end{table*}

\begin{table*}[t]
\centering
\scriptsize
\setlength{\tabcolsep}{3.5pt}
\renewcommand{\arraystretch}{1.05}
\resizebox{\textwidth}{!}{%
\begin{tabular}{lrrrrrrrrr}
\toprule
\textbf{Model} &
\textbf{Usable} &
\textbf{H-DI} &
\textbf{H-MI\c{S}} &
\textbf{L-DI} &
\textbf{L-MI\c{S}} &
\textbf{DI\% (H)} &
\textbf{DI\% (L)} &
\textbf{$\Delta$DI (pp)} &
\textbf{$p$} \\
\midrule
Gemma-3-27B-IT        & 96.0 & 93 & 2  & 82 & 15 & 97.9 & 84.5 & +13.4 & $1.51\times10^{-3}$ \\
Gemma-2-9B-IT-TR      & 91.0 & 92 & 0  & 88 & 2  & 100.0 & 97.8 & +2.2  & $0.243$ \\
Orbita-v0.1           & 22.0 & 19 & 6  & 11 & 8  & 76.0 & 57.9 & +18.1 & $0.327$ \\
Qwen3-32B             & 32.5 & 24 & 3  & 35 & 3  & 88.9 & 92.1 & -3.2  & $0.686$ \\
Novus-7B-TR           & 28.0 & 27 & 6  & 21 & 2  & 81.8 & 91.3 & -9.5  & $0.449$ \\
WiroAI-TR-8B          & 24.5 & 24 & 3  & 20 & 2  & 88.9 & 90.9 & -2.0  & $1.000$ \\
GPT-OSS-20B-Astro     & 27.5 & 17 & 11 & 19 & 8  & 60.7 & 70.4 & -9.7  & $0.573$ \\
GPT2-TR-10M           & 4.5  & 3  & 0  & 6  & 0  & 100.0 & 100.0 & +0.0  & -- \\
Turkcell-LLM-7B       & 2.0  & 3  & 0  & 1  & 0  & 100.0 & 100.0 & +0.0  & -- \\
Trendyol-LLM-7B       & 0.0  & 0  & 0  & 0  & 0  & --    & --    & --    & -- \\
\bottomrule
\end{tabular}%
}
\caption{Experiment~II (explicit past-tense generation) LLM results. Each item was generated three times (\texttt{tur1--tur3}), and the final item-level label was determined by majority vote among outputs classifiable as \textsc{DI} or \textsc{mI\c{s}} (ties excluded). \textbf{Usable} = percentage of all items (out of 200) receiving a majority \textsc{DI} or \textsc{mI\c{s}} label. DI\% is computed as $\mathrm{DI}/(\mathrm{DI}+\mathrm{MIS})$ within each trust condition. $\Delta$DI is High minus Low (percentage points). $p$ values are Fisher exact tests on \{High, Low\}$\times$\{\textsc{DI}, \textsc{mI\c{s}}\} using item-level majority labels.}
\label{tab:llm-exp2-main}
\end{table*}

\section{The First Experiment with LLMs} 
Experiment~I evaluates LLM behavior in a prompting-based gap-fill task using 200 trust-manipulated Turkish cloze items. Each item was presented to the model three times, and the final label was determined by majority vote across the three outputs. Table~\ref{tab:llm-exp1-main}
summarizes the model-level results on the prompted task, reporting (i) the
percentage of \emph{usable} outputs (i.e., generations classifiable as
\textsc{DI} or \textsc{mI\c{s}}), (ii) run-level \textsc{DI}/\textsc{mI\c{s}}
counts in High- and Low-Trust conditions, (iii) the within-condition
\textsc{DI} share, and (iv) Fisher exact tests for the High-vs.-Low trust
contrast. A first important result is that the amount of usable evidential data
varies substantially across models, ranging from 64.5\% (Gemma-3-27B-IT) to
0.0\% (Trendyol-LLM-7B). This means that, in the prompting setup, a large
portion of outputs may be off-target and therefore cannot be interpreted as an
evidential choice.

Among models with substantial usable \textsc{DI}/\textsc{mI\c{s}} outputs, the
trust effect is not uniform. Gemma-3-27B-IT shows a clear and statistically
reliable shift in the expected direction: the \textsc{DI} share decreases from
98.5\% in High-Trust contexts to 88.9\% in Low-Trust contexts
($\Delta$DI $=+9.6$ pp, $p=7.44\times10^{-5}$), indicating relatively more
\textsc{mI\c{s}} in Low-Trust items. Gemma-2-9B-IT-TR shows a similar but
smaller significant pattern (98.8\% vs.\ 91.9\%; $\Delta$DI $=+6.9$ pp,
$p=0.006$). In contrast, Orbita-v0.1 shows a significant effect in the
\emph{opposite} direction, with a higher \textsc{DI} share in Low-Trust than in
High-Trust contexts (81.5\% vs.\ 67.5\%; $\Delta$DI $=-14.0$ pp, $p=0.018$).

The remaining models mostly show null, unstable, or weak effects. Qwen3-32B,
Novus-7B-TR, and WiroAI-TR-8B exhibit small and non-significant differences
($p=.371$, $.556$, and $.696$, respectively). GPT-OSS-20B-Astro shows a larger
numerical reverse shift ($\Delta$DI $=-20.4$ pp) but does not reach
conventional significance ($p=.058$). For GPT2-TR-10M and Turkcell-LLM-7B, all
usable outputs are \textsc{DI}, so no informative \textsc{DI}/\textsc{mI\c{s}}
comparison can be made; Trendyol-LLM-7B yields no usable outputs. Overall,
Experiment~I suggests that in an open-generation prompting setup, trust-sensitive
evidential behavior is highly model-dependent and often overshadowed by
prompting/compliance limitations and a strong general bias toward \textsc{DI}.


\section{The Second Experiment with LLMs}
\label{sec:llm-exp2}

Experiment~II evaluates LLM behavior in an \emph{explicit past-tense}
generation setup using the same 200 trust-manipulated Turkish cloze items.
Unlike Experiment~I, the prompt here explicitly instructed models to produce a
\emph{past-tense} completion. As in Experiment~I, each item was sampled three
times. To obtain a single interpretable evidential label per item, we assigned a
final item-level label by majority vote across the three outputs
(\texttt{tur1--tur3}) among outputs classifiable as \textsc{DI} or
\textsc{mI\c{s}} (ties excluded).

Table~\ref{tab:llm-exp2-main} summarizes the model-level results for the
explicit-past prompting condition, reporting (i) the percentage of
\emph{usable} items (i.e., items receiving a majority label classifiable as
\textsc{DI} or \textsc{mI\c{s}}), (ii) item-level \textsc{DI}/\textsc{mI\c{s}}
counts in High- and Low-Trust conditions, (iii) the within-condition
\textsc{DI} share, and (iv) Fisher exact tests for the High-vs.-Low trust
contrast. A first key result is that usability remains strongly
model-dependent even with explicit past-tense instructions, ranging from 96.0\%
(Gemma-3-27B-IT) and 91.0\% (Gemma-2-9B-IT-TR) to 0.0\%
(Trendyol-LLM-7B). Thus, explicit past-tense prompting substantially improves
the proportion of usable evidential outputs for some models, but not for all.

Among models with substantial usable \textsc{DI}/\textsc{mI\c{s}} outputs,
Gemma-3-27B-IT shows the clearest and statistically reliable trust-sensitive
pattern in the predicted direction. Its \textsc{DI} share decreases from 97.9\%
in High-Trust contexts to 84.5\% in Low-Trust contexts
($\Delta$DI $=+13.4$ pp, $p=1.51\times10^{-3}$), while the number of
\textsc{mI\c{s}} labels increases from 2 items (High) to 15 items (Low). This
indicates a relative increase in \textsc{mI\c{s}} under Low-Trust contexts.
Gemma-2-9B-IT-TR also shows high usability and a small numerical shift in the
expected direction (100.0\% vs.\ 97.8\%; $\Delta$DI $=+2.2$ pp), but the
difference is not statistically significant ($p=.243$). Orbita-v0.1 shows a
larger numerical shift in the expected direction ($\Delta$DI $=+18.1$ pp), but
its low usability (22.0\%) and non-significant test ($p=.327$) limit the
strength of this result.

The remaining models mostly show null, unstable, or reverse-direction patterns.
Qwen3-32B, Novus-7B-TR, WiroAI-TR-8B, and GPT-OSS-20B-Astro all show
non-significant High-vs.-Low differences ($p=.686$, $.449$, $1.000$, and
$.573$, respectively), and several exhibit a numerical reverse shift (i.e., a
higher \textsc{DI} share in Low-Trust than in High-Trust contexts). For
GPT2-TR-10M and Turkcell-LLM-7B, all usable majority labels are \textsc{DI}, so
no informative \textsc{DI}/\textsc{mI\c{s}} comparison can be made. Trendyol-LLM-7B
produces no usable items. Overall, Experiment~II shows that explicit
past-tense prompting can improve evidential compliance and reveal a
trust-sensitive \textsc{DI}/\textsc{mI\c{s}} pattern in some models (most
clearly Gemma-3-27B-IT), but the effect remains highly model-dependent and is
weak, unstable, or absent in most others.

\subsection{Exploratory Head-Level Modulation of Trust-Cue Attention}

As an exploratory mechanistic analysis, we examined attention-weight redistribution in \textit{Gemma-2-9B-IT-TR} under the explicit past-tense prompting condition. We focused on the pre-generation decision state (\texttt{q = last\_prompt}) and quantified attention directed to the trust-cue span, operationalized as the token sequence immediately preceding \textit{g\"ore}. Across heads, the strongest condition sensitivity was confined to a small subset of mid-to-late layers, with the single best layer at $L=25$ and the highest-scoring layers concentrated in the range $L=22$--$29$. However, the aggregate pairwise contrast was weak and directionally unstable ($\Delta_{\text{pair}}$ mean $=-0.0179$, 95\% bootstrap CI $[-0.0513,\,0.0170]$; $P(\Delta_{\text{pair}}>0)=46\%$). Under this operationalization, the results do not provide strong evidence that trust is robustly encoded via systematic reallocation of attention mass to the cue span in \textit{Gemma-2-9B-IT-TR}; accordingly, we treat the attention maps as qualitative diagnostic evidence rather than as a stable mechanistic signature. The corresponding visualizations are provided in Appendix ~\ref{app:Exploratory Attention Maps}.

\begin{table*}[t]
\centering
\scriptsize
\setlength{\tabcolsep}{3.5pt}
\renewcommand{\arraystretch}{1.05}
\resizebox{\textwidth}{!}{%
\begin{tabular}{lrrrrrrrr}
\toprule
\textbf{Model} &
\textbf{DI\% High} &
\textbf{DI\% Low} &
\textbf{$\Delta$pp} &
\textbf{Acc\%} &
\textbf{Abstain\%} &
\textbf{Fisher $p$} &
\textbf{Holm $p$} &
\textbf{3-run same\%} \\
\midrule
Gemma-2-9B-IT-TR   & 60.0 & 44.0 & 16.0 & 58.0 & 0.0  & 0.033 & 0.268 & 97.5 \\
Gemma-3-27B-IT     & 78.9 & 66.3 & 12.6 & 56.0 & 3.5  & 0.054 & 0.381 & 97.5 \\
GPT-OSS-20B-Astro  & 94.9 & 84.0 & 10.9 & 55.3 & 0.5  & 0.019 & 0.173 & 92.0 \\
Orbita-v0.1        & 91.0 & 84.0 & 7.0  & 53.5 & 0.0  & 0.199 & 0.994 & 96.0 \\
Turkcell-LLM-7B    & 97.0 & 90.0 & 7.0  & 53.3 & 0.5  & 0.082 & 0.491 & 81.0 \\
WiroAI-TR-8B       & 15.2 & 11.0 & 4.2  & 52.3 & 0.5  & 0.408 & 1.000 & 91.0 \\
Qwen3-32B          & 81.0 & 77.0 & 4.0  & 52.0 & 0.0  & 0.603 & 1.000 & 95.0 \\
Novus-7B-TR        & 99.0 & 98.0 & 1.0  & 50.5 & 0.0  & 1.000 & 1.000 & 99.0 \\
Trendyol-LLM-7B    & 0.0  & 0.0  & 0.0  & 50.0 & 0.0  & 1.000 & 1.000 & 100.0 \\
GPT2-TR-10M        & 0.0  & --   & --   & 0.0  & 99.5 & --    & --    & 99.5 \\
\bottomrule
\end{tabular}%
}
\caption{Experiment~III (A/B forced-choice prompting) combined model results using item-level majority vote over three runs. DI\% High/Low are computed after excluding abstentions (invalid outputs and tie/no-majority items). $\Delta$pp = DI\% High minus DI\% Low. Acc\% uses the target mapping High$\rightarrow$\textit{-DI}, Low$\rightarrow$\textit{-mI\c{s}}. Fisher exact tests are computed on High/Low $\times$ \{\textsc{DI}, \textsc{mI\c{s}}\} after excluding abstentions; Holm $p$ values correct for multiple testing across models. 3-run same\% is the proportion of items with identical outputs across all three replicates.}
\label{tab:llm-exp3-main}
\end{table*}

Applying the same analysis to \textit{Gemma-3-27B-IT} yielded a larger and more internally consistent pair-level contrast. Condition-sensitive modulation was again localized to a restricted subset of mid-to-late-layer heads, but with a clearer concentration in higher layers (best layer $L=31$; top layers $L=31,33,36,37$). The aggregate pairwise effect was reliably non-zero under the present metric definition ($\Delta_{\text{pair}}$ mean $=-0.0983$, 95\% bootstrap CI $[-0.1151,\,-0.0817]$; $P(\Delta_{\text{pair}}>0)=9\%$), indicating systematic condition-dependent redistribution of attention toward the trust-cue span. We interpret this as evidence that \textit{Gemma-3-27B-IT} exhibits a more coherent internal sensitivity to trust-related contextual cues than \textit{Gemma-2-9B-IT-TR} in this prompting setup. At the same time, this interpretation should remain qualified, since raw span-level attention mass can be affected by surface differences such as cue-length asymmetries across conditions. The relevant maps are shown in Appendix ~\ref{app:Exploratory Attention Maps}.


\section{The Third Experiment with LLMs}
\label{sec:llm-exp3}

Experiment~III evaluates LLM behavior in a \emph{forced-choice} prompting
setup, in which models are explicitly asked to choose between two candidate
completions: a \DI{} form (option A) and a \MIS{} form (option B). This design
is more constrained than Experiments~I--II because it removes open-ended lexical
generation and directly targets the evidential choice itself. The same 200
trust-manipulated Turkish items are used (100 High-Trust, 100 Low-Trust), and
each item is queried three times (\texttt{tur1--tur3}). Final item-level labels
are determined by majority vote across the three runs. Outputs other than
\texttt{A}/\texttt{B}, as well as ties/no-majority cases, are treated as
abstentions and excluded from evidential-rate and Fisher-test calculations.

Table~\ref{tab:llm-exp3-main} reports per-model condition-wise \textsc{DI} rates
(\textsc{DI}\% High and \textsc{DI}\% Low), the High-minus-Low difference
($\Delta$pp), target-mapping accuracy (High$\rightarrow$\DI{}, Low$\rightarrow$\MIS{}),
and abstention rate over all 200 items. Table~\ref{tab:llm-exp3-main} reports
per-model Fisher exact tests (High/Low $\times$ \textsc{DI}/\textsc{MIS}),
Holm-corrected $p$ values across models, and 3-run agreement (\emph{same}\%),
which measures response stability across the three replicates.

A first important result is that most usable models show a
\emph{directionally trust-consistent trend}: \textsc{DI}\% is higher in
High-Trust than in Low-Trust contexts. The largest shift is observed for
\model{neuralwork/gemma-2-9b-it-tr} ($\Delta$pp $=16.0$), followed by
\model{google/gemma-3-27b-it} ($12.6$ pp) and
\model{ocaklisemih/gpt-oss-20b-turkish-astrology-gguf} ($10.9$ pp). Several
other models also show smaller positive shifts (e.g.,
\model{mradermacher/Orbita-v0.1-i1-GGUF}, \model{Turkcell-LLM-7b-v1},
\model{WiroAI/wiroai-turkish-llm-8b}, \model{Qwen/Qwen3-32B},
\model{NovusResearch/Novus-7b-tr_v1}).

\model{Trendyol-LLM-7b-base-v1.0} shows no shift (0.0 vs.\ 0.0). This pattern
suggests that, under direct A/B comparison, many models weakly track the trust
manipulation in the predicted direction.

At the same time, the results also show that \emph{base-rate preferences remain
strong} and often dominate behavior. Several models overwhelmingly prefer one
option across both conditions (e.g., very high \textsc{DI} rates for Novus,
Turkcell, Orbita, and GPT-OSS; near-zero \textsc{DI} rates for Trendyol and
GPT2-TR-10M), which compresses the effect of trust and limits target-mapping
accuracy. Consistent with this, Acc\% values remain close to chance for most
models (roughly 50--58\%), even when the direction of the trust effect is
correct. In other words, models can show a small High$>$Low \textsc{DI} shift
without demonstrating robust context-sensitive evidential selection at the item
level.

Statistical testing reinforces this conclusion. Some models yield nominally
small Fisher exact $p$ values (e.g., $p=.019$ for GPT-OSS and $p=.033$ for
Gemma-2-9B-IT-TR; Table~\ref{tab:llm-exp3-main}), but \emph{none} remains
significant after Holm correction across models. Thus, Experiment~III provides
evidence for a weak, directionally consistent trend in several systems, but not
for a robust per-model trust effect after correction for multiple comparisons.

Finally, response stability across the three replicates is generally high
(3-run same\% often above 90\%), indicating that model choices in this A/B
paradigm are relatively deterministic once the prompt format is fixed. This is
methodologically useful: compared with open-ended prompting, the forced-choice
setup reduces output-format variability and improves comparability across models.
However, the combination of high replicate agreement and near-chance Acc\% for
many systems suggests that the models are often making stable choices driven by
global option preferences rather than reliably applying the trust cue in a
human-like way.

\section{Discussion}
\label{sec:discussion}

This study makes theoretical and methodological contributions to Turkish evidentiality and its evaluation in LLMs. Human results show a clear trust effect: more credible external sources favor \textit{-DI}, while less credible sources favor \textit{-mI\c{s}} (Section~\ref{sec:human-results}; Table~\ref{tab:strict-counts}; Figure~\ref{fig:di-mis-by-trust}). In contrast, current LLMs do not reliably reproduce this pattern, and their apparent evidential behavior varies with prompt format, output compliance, and response constraints (Tables~\ref{tab:llm-exp1-main}, \ref{tab:llm-exp2-main}, \ref{tab:llm-exp3-main}).

The paper’s central theoretical contribution is a direct experimental test of a \emph{trustworthiness}-based perspective on Turkish evidentiality, inspired by the Giannakidou--Mari framework. To our knowledge, this is a novel application of that framework to Turkish \textit{-DI}/\textit{-mI\c{s}}. Rather than treating evidential morphology only as a categorical source-type marker, we test whether the credibility of an explicitly external source shifts evidential choice; the human data show that it does. 

This matters because the source is explicitly external in both conditions ((e.g., X’e göre ‘according to X’)); the manipulation changes only the source’s trustworthiness. The observed shift therefore supports an analysis in which Turkish evidential morphology is sensitive not just to source \emph{type}, but also to source \emph{quality} and the speaker’s resulting commitment profile. In Giannakidou--Mari terms, the pattern is consistent with differences in support strength for the prejacent: more trustworthy sources favor stronger commitment (and relatively more \textit{-DI}), whereas less trustworthy sources favor weaker, evidence-weighted commitment (and relatively more \textit{-mI\c{s}}) \citep{giannakidoumari2016,giannakidoumari2018,giannakidoumari2021,boscaro-giannakidou-mari-2025}.

Our findings do not by themselves settle the long-standing question of whether \textit{-DI} is lexically a direct evidential or a past tense form whose ``directness'' effects arise pragmatically. However, they do provide an important constraint on the debate. Because the manipulation keeps source attribution explicit and varies trustworthiness instead, the human pattern is difficult to reduce to a simple source-present/source-absent contrast. The results are more naturally captured by accounts that allow context-sensitive interactions among evidential morphology, speaker commitment, and discourse-pragmatic reasoning. In this sense, the present data strengthen the case for treating Turkish evidential choice as part of a broader inferential and commitment-sensitive system, rather than a purely rigid source-labeling mechanism.

\paragraph{What the LLM results may suggest.}
Across the three LLM experiments, the main result is a dissociation between \emph{surface task performance under prompting constraints} and \emph{human-like trust-sensitive evidential reasoning}. In Experiment~I (open gap-fill prompting), many models produce a high proportion of unusable outputs, and even among usable outputs the trust effect is highly model-dependent. In Experiment~II (explicit past-tense prompting), usability improves substantially for some models (especially Gemma variants), and Gemma-3-27B-IT shows the clearest trust-consistent shift, but the pattern remains weak, absent, or unstable in most models. In Experiment~III (A/B forced-choice), response stability becomes high and several models show directionally trust-consistent trends, yet effects are generally modest, target-mapping accuracy remains near chance for many systems, and no per-model effect survives Holm correction. Thus, these results suggest that many LLMs can exhibit \emph{format-dependent traces} of the expected pattern, but they do not reliably reproduce the human sensitivity to source trustworthiness as a stable, cross-paradigm property. This is precisely the kind of mismatch that a linguistically targeted, human--LLM comparison can reveal: a model may appear ``reasonable'' in one prompt format while still failing to encode the underlying contextual factor in a robust way. A related question is why explicit past-tense prompting changes model performance. We suggest that the main reason is task specification: in the open gap-fill setting, models must jointly infer a plausible lexical completion, tense, and evidential morphology, which leaves the task relatively underspecified. By explicitly requiring a past-tense completion in Experiment~II, the prompt narrows the response space and makes the \textit{-DI}/\textit{-mI\c{s}} contrast more directly relevant to the decision. In this sense, the improvement for some models is consistent with recent work showing that more specific instructions can reduce prompt sensitivity relative to underspecified prompts, even if they do not solve the deeper representational problem \citep{pecher-etal-2026-prompt}. At the same time, our results also align with recent findings that morphological generalization remains difficult for LLMs in agglutinative languages such as Turkish, especially when models must produce the appropriate inflection rather than merely recognize it \citep{ismayilzada-etal-2025-morphology}.

The paper also contributes a reusable evaluation paradigm for Turkish source-sensitive semantics/pragmatics by combining: (i) a human cloze dataset, (ii) an expanded LLM-only dataset, and (iii) three prompting regimes that vary constraint level (open generation, explicit tense generation, forced choice). This multi-paradigm design makes it possible to separate at least three sources of variation that are often conflated in LLM evaluations: \emph{compliance/formatting behavior}, \emph{base-rate morphological preference}, and \emph{context sensitivity to the trust manipulation}. In particular, the contrast between Experiments~I--III shows that forcing the response space can improve comparability and stability, but it can also expose strong default biases that limit genuine contextual adaptation. More broadly, the study argues for evaluating Turkish LLMs on controlled linguistic contrasts with human baselines, rather than relying only on broad benchmarks. A useful point of comparison is \textit{TurBLiMP}, the first Turkish benchmark of linguistic minimal pairs, which evaluates LMs on 16 grammatical phenomena and has significantly expanded linguistically informed evaluation for Turkish \citep{basar-etal-2025-turblimp}. Our study is complementary to that line of work. Whereas TurBLiMP focuses on minimal-pair judgments over a broad range of grammatical phenomena, it does not target evidentiality. This matters because evidentiality is not only a morphosyntactic contrast in Turkish, but also a source-sensitive semantic--pragmatic phenomenon tied to speaker commitment, information source, and discourse reasoning. Our results therefore extend Turkish LLM evaluation into a domain that is central to Turkish grammar but not captured by existing minimal-pair benchmarks. For evidentiality and related phenomena, this is especially important because the relevant behavior is not merely lexical accuracy but context-dependent mapping between linguistic form and epistemic/discourse structure.

\section{Conclusion}
\label{sec:conclusion}

This paper investigated how \emph{source trustworthiness} shapes Turkish evidential choice in the \textit{-DI}/\textit{-mI\c{s}} contrast, and whether current LLMs track this sensitivity. In a controlled human cloze experiment, we found a robust and replicable trust effect: participants produced relatively more \textit{-DI} in High-Trust contexts and relatively more \textit{-mI\c{s}} in Low-Trust contexts. This provides new empirical support for a trust-/commitment-based perspective on Turkish evidentiality and shows that the Giannakidou--Mari framework offers a productive theoretical lens for Turkish data.

Across three LLM experiments (open gap-fill prompting, explicit past-tense prompting, and forced-choice A/B prompting), model behavior was substantially more variable and strongly dependent on prompt format, output compliance, and base-rate suffix preferences. While some models showed weak or local trust-consistent trends, the effect was not robust across models or paradigms. Overall, the results reveal a clear human--LLM gap in source-sensitive evidential reasoning and motivate future work on more faithful evaluation and modeling of evidential and commitment-related meaning in Turkish.

\section*{Ethics Statement}
This work investigated the interaction between contextual source trustworthiness and Turkish evidential morphology (\textit{-DI} vs.\ \textit{-mI\c{s}}) using both human cloze responses and LLM-based preference scoring. Human data were collected under informed-consent procedures and analyzed only at the aggregate level; no personally identifying information is reported, and all examples are presented in anonymized form. The human-subjects component of this study received Institutional Review Board (IRB) approval.

For the computational component, model outputs are treated as task-conditioned behavioral responses rather than as evidence of human-like semantic or cognitive representations. Because LLM preferences can be sensitive to evaluation protocol design (including prompting format and tokenization effects), our findings should not be overgeneralized to broad claims about model competence or human evidential reasoning. Finally, by focusing on Turkish, this study helps address the English-centric bias of much computational psycholinguistics and LLM evaluation.

\section*{Limitations}
\label{app:limitations}

This study has several limitations. First, the strongest behavioral evidence comes from the human experiment on Dataset~1, whereas the larger Dataset~2 was used only for LLM evaluation; future work should test whether the same human trust effect scales to the expanded item set. Second, LLM results are partly constrained by prompt compliance and output-format errors, especially in open-generation settings, which can reduce the amount of usable evidential data and make model comparisons harder to interpret. Third, although the forced-choice setup improves control and stability, it may introduce response biases (e.g., option or format preferences) that are not specific to evidential reasoning. The exploratory attention analysis is also limited in scope since it was conducted only for two relatively successful models from Experiment~II and is intended as a preliminary qualitative diagnostic rather than a definitive mechanistic account.

\section*{Acknowledgements}
We thank Anastasia Giannakidou for valuable input on the semantic and pragmatic theory and the interpretation of the results, Ming Xiang for suggestions on the human experimental design, and Christopher Potts and Craig Thorburn for their comments on future directions. We are also grateful to the anonymous reviewers for their helpful feedback and suggestions. Any remaining errors are our own.

\section{Bibliographical References}\label{sec:reference}

\bibliographystyle{lrec2026-natbib}
\bibliography{lrec2026-example}



\clearpage
\appendix

\clearpage
\section{Source Categories}
\label{app:source_categories}

\begin{table}[h]
\centering
\small
\rowcolors{2}{gray!10}{white}
\begin{tabular}{>{\centering\arraybackslash}p{2.5cm} p{5.5cm} p{6cm}}
\toprule
\rowcolor{gray!25}
\textbf{Condition} & \textbf{Turkish} & \textbf{English} \\
\midrule
High & Resmi Gazete & Official Gazette \\
High & Kurumun resmi sayfası & Official website of the institution \\
High & Hastanenin astığı duyuru & Hospital notice board announcement \\
High & Belediyenin SMS uyarısı & Municipality SMS alert \\
High & Resmi mail & Official email \\
\midrule
Low & Anonim telegram grubu & Anonymous Telegram group \\
Low & Yan apartmanda oturan komşu & Neighbor living in the adjacent building \\
Low & Diğer şirketteki çalışan & Employee at another company \\
Low & Yanlış haberler paylaşan Instagram grubu & Instagram group sharing false news \\
Low & Sokakta geçen biri & A random passerby on the street \\
\bottomrule
\end{tabular}
\caption{Examples of high- and low-trust source categories used in the experiment.}
\end{table}

\clearpage
\section{Prompt Templates}
\label{app:prompts}

\tcbset{colback=gray!5, colframe=black!40, boxrule=0.6pt, arc=2mm}

\subsection{Turkish Prompts}
\label{app:prompts-tr}

\begin{enumerate}[leftmargin=*, label=\textbf{Prompt \arabic*:}, itemsep=10pt]

\item \label{app:prompt1}
\begin{tcolorbox}[breakable, title={Prompt 1 (TR)}]
\small
\raggedright
A\c{s}a\u{g}\i daki c\"umlede ``\_\_\_\_\_\_\_\_\_\_.'' ile belirtilmi\c{s} bir bo\c{s}luk var.\par
Bu bo\c{s}lu\u{g}u c\"umlenin yap\i s\i n\i n ak\i \c{s}\i na en uygun fiille tamamla.\par

\medskip
Cevap verirken kurallar:\par
- Sadece bo\c{s}lu\u{g}a gelecek k\i sm\i  yaz.\par
- C\"umleyi tekrar etme.\par
- A\c{c}\i klama yapma.\par
Tek kelime cevap ver ve fiil kullanarak tamamla\par
- Cevap vermeden ge\c{c}me.\par

\medskip
C\"umle:\par
\texttt{\{prefix\}} \_\_\_\_\_\_\_\_\_\_.\par

\medskip
Cevap:
\end{tcolorbox}

\item \label{app:prompt2}
\begin{tcolorbox}[breakable, title={Prompt 2 (TR)}]
\small
\raggedright
A\c{s}a\u{g}\i daki c\"umlede ``\_\_\_\_\_\_\_\_\_\_.'' ile belirtilmi\c{s} bir bo\c{s}luk var.\par
Bu bo\c{s}lu\u{g}u c\"umlenin yap\i s\i n\i n ak\i \c{s}\i na ve ge\c{c}mi\c{s} zamana g\"ore tamamlanabilecek en uygun kelimeyle tamamla.\par

\medskip
Cevap verirken kurallar:\par
- Sadece bo\c{s}lu\u{g}a gelecek k\i sm\i  yaz.\par
- C\"umleyi tekrar etme.\par
- A\c{c}\i klama yapma.\par
- Cevap vermeden ge\c{c}me.\par

\medskip
C\"umle:\par
\texttt{\{prefix\}} \_\_\_\_\_\_\_\_\_\_.\par

\medskip
Cevap:
\end{tcolorbox}

\item \label{app:prompt3}
\begin{tcolorbox}[breakable, title={Prompt 3 (TR)}]
\small
\raggedright
C\"umledeki bo\c{s}lu\u{g}u doldurmak i\c{c}in do\u{g}ru se\c{c}ene\u{g}i se\c{c}.\par

\medskip
C\"umle: \texttt{\{sentence\}}\par

\medskip
Se\c{c}enekler:\par
A) \texttt{\{di\}}\par
B) \texttt{\{mis\}}\par

\medskip
Kurallar:\par
- Sadece A ya da B yaz.\par
- A\c{c}\i klama yapma.\par

\medskip
Cevap:
\end{tcolorbox}

\end{enumerate}

\subsection{English Prompts}
\label{app:prompts-en}

\begin{enumerate}[leftmargin=*, label=\textbf{Prompt \arabic*:}, itemsep=10pt]

\item
\begin{tcolorbox}[breakable, title={Prompt 1 (EN)}]
\small
\raggedright
There is a blank indicated by ``\_\_\_\_\_\_\_\_\_\_.'' in the sentence below.\par
Fill this blank with the verb that best fits the flow and structure of the sentence.\par

\medskip
Rules for your answer:\par
- Write only what should go in the blank.\par
- Do not repeat the sentence.\par
- Do not provide an explanation.\par
Answer with a single word and complete it using a verb.\par
- Do not skip answering.\par

\medskip
Sentence:\par
\texttt{\{prefix\}} \_\_\_\_\_\_\_\_\_\_.\par

\medskip
Answer:
\end{tcolorbox}

\item
\begin{tcolorbox}[breakable, title={Prompt 2 (EN)}]
\small
\raggedright
There is a blank indicated by ``\_\_\_\_\_\_\_\_\_\_.'' in the sentence below.\par
Fill this blank with the most appropriate word that can complete the sentence according to the flow/structure and past tense.\par

\medskip
Rules for your answer:\par
- Write only what should go in the blank.\par
- Do not repeat the sentence.\par
- Do not provide an explanation.\par
- Do not skip answering.\par

\medskip
Sentence:\par
\texttt{\{prefix\}} \_\_\_\_\_\_\_\_\_\_.\par

\medskip
Answer:
\end{tcolorbox}

\item
\begin{tcolorbox}[breakable, title={Prompt 3 (EN)}]
\small
\raggedright
Choose the correct option to fill the blank in the sentence.\par

\medskip
Sentence: \texttt{\{sentence\}}\par

\medskip
Options:\par
A) \texttt{\{di\}}\par
B) \texttt{\{mis\}}\par

\medskip
Rules:\par
- Write only A or B.\par
- Do not provide an explanation.\par

\medskip
Answer:
\end{tcolorbox}

\end{enumerate}

\clearpage
\section{Exploratory Attention Maps}
\label{app:Exploratory Attention Maps}

\begin{figure}[!ht]
\centering
\includegraphics[width=\columnwidth]{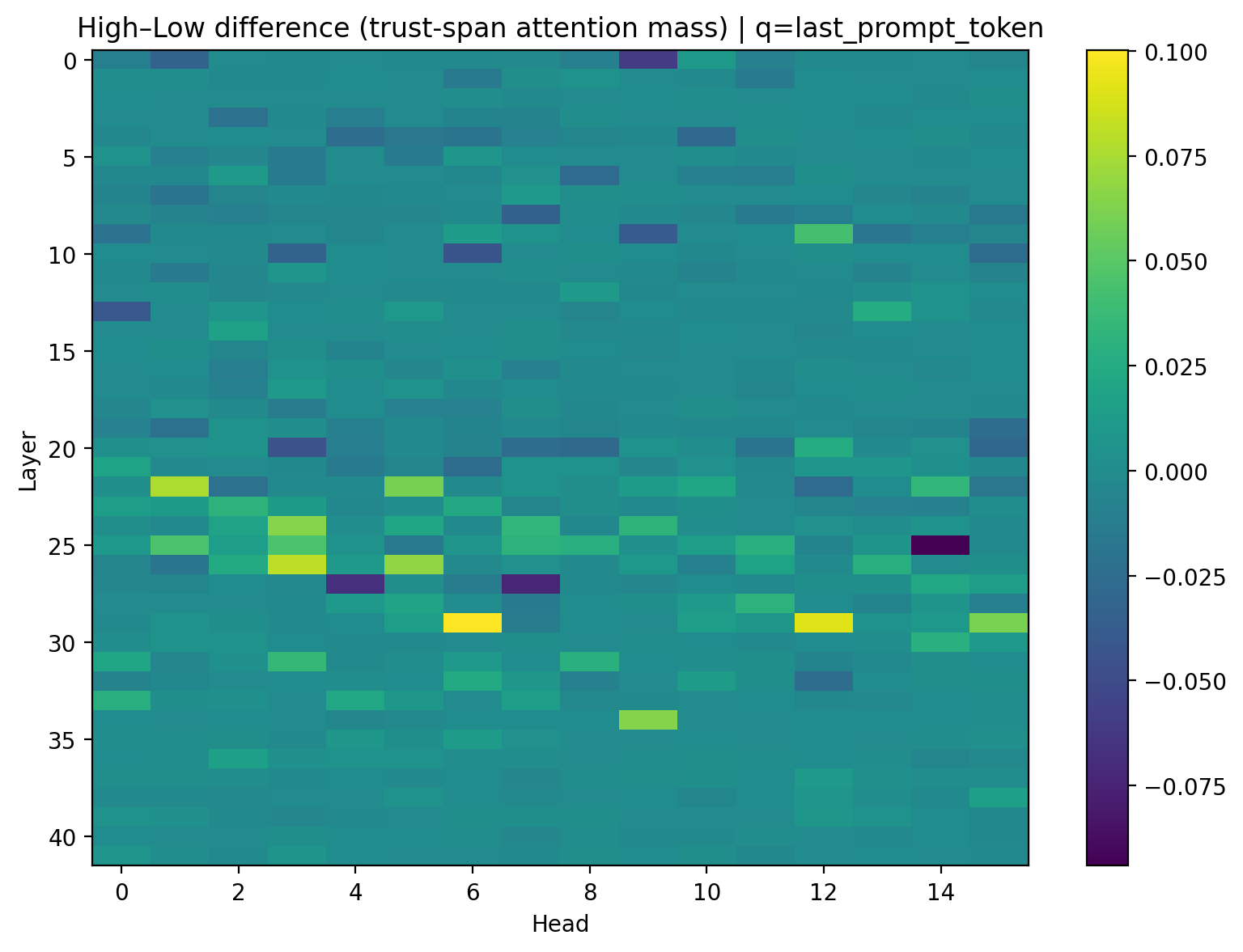}
\caption{Gemma-2-9B-IT-TR (Experiment~II): high--low difference in trust-span attention mass (query token: \texttt{last\_prompt\_token}). The pair-level effect is weak and inconsistent ($\Delta_{\text{pair}}$ mean $=-0.0179$, 95\% bootstrap CI $[-0.0513,\,0.0170]$; $P(\Delta_{\text{pair}}>0)=46\%$), so we treat this as a qualitative diagnostic rather than strong evidence of robust trust encoding.}
\label{fig:gemma2-attn-highlow}
\end{figure}

\FloatBarrier

\begin{figure}[!ht]
\centering
\includegraphics[width=\columnwidth]{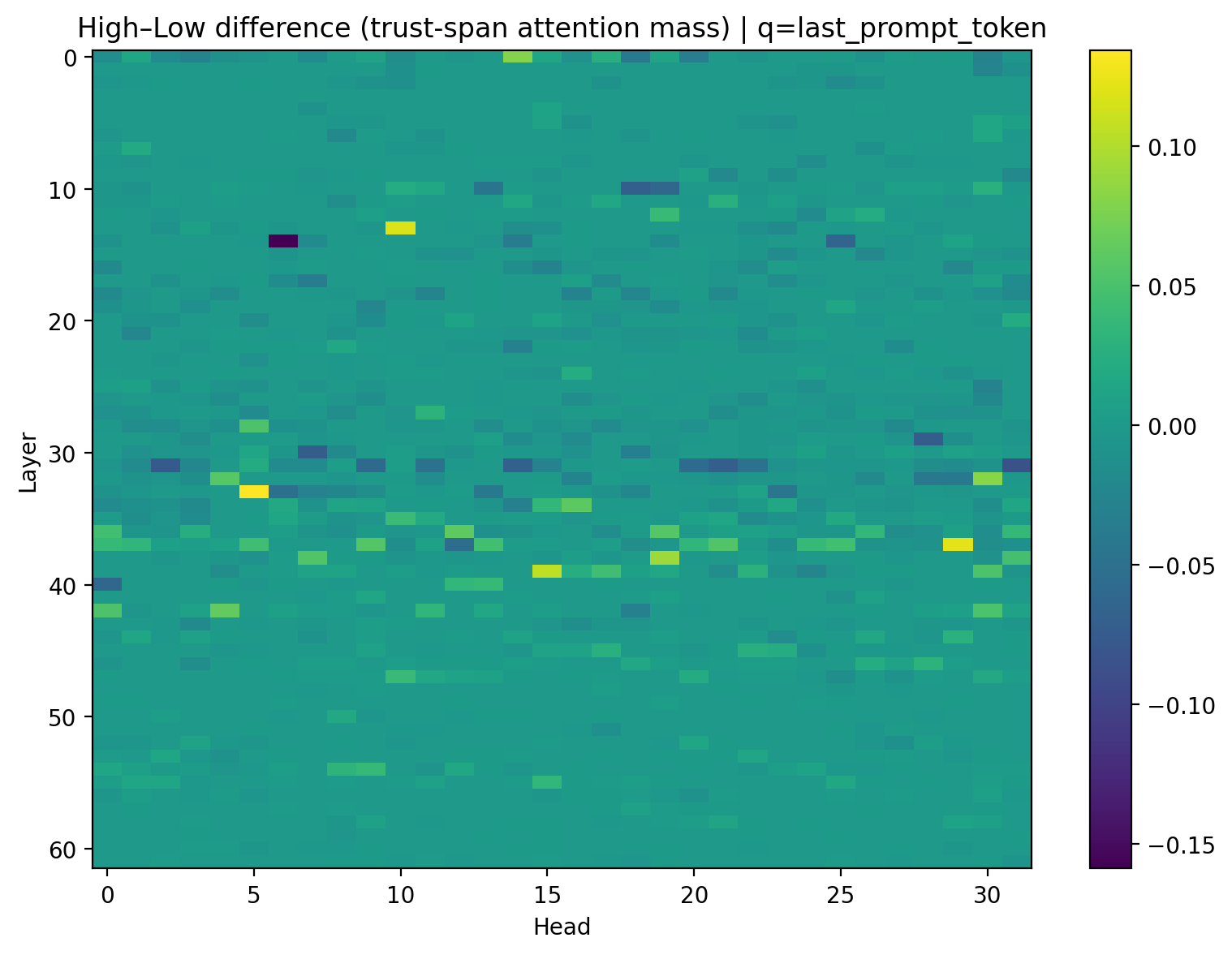}
\caption{Gemma-3-27B-IT (Experiment~II): high--low difference in trust-span attention mass (query token: \texttt{last\_prompt\_token}). The map shows localized modulation concentrated in a small cluster of mid-to-late layers and heads, consistent with the exploratory analysis in Section~\ref{sec:llm-exp2}.}
\label{fig:gemma3-attn-highlow}
\end{figure}

\end{document}